\documentclass[10pt,twocolumn,letterpaper]{article}

\usepackage{cvpr}
\usepackage{times}
\usepackage{epsfig}
\usepackage{graphicx}
\usepackage{amsmath}
\usepackage{bm}
\usepackage{amssymb}
\usepackage{mathabx}

\usepackage{mathrsfs}
\usepackage{caption}
\usepackage{pifont}
\usepackage{wasysym}
\usepackage{array}
\usepackage{caption}
\usepackage{colortbl}
\usepackage{xcolor}
\newcommand*\colourcheck[1]{%
  \expandafter\newcommand\csname #1check\endcsname{\textcolor{#1}{\ding{52}}}%
}
\colourcheck{red}
\newcommand{\xmark}{\ding{55}}%
\DeclareMathOperator{\E}{\mathbb{E}}
\usepackage[pagebackref=true,breaklinks=true,letterpaper=true,colorlinks,bookmarks=false]{hyperref}

\cvprfinalcopy 

\def\cvprPaperID{7715} 

\ifcvprfinal\fi
\begin{document}

\title{EventSR: From Asynchronous Events to Image Reconstruction, Restoration, and Super-Resolution via End-to-End Adversarial Learning}

\author{Lin Wang$^{1}$, Tae-Kyun Kim$^{2}$,
and Kuk-Jin Yoon$^{1}$\\
$^{1}$Visual Intelligence Lab., KAIST, Korea\\
$^{2}$ICVL Lab., Imperial College London, UK \\
{\tt\small wanglin@kaist.ac.kr, tk.kim@imperial.ac.uk,
kjyoon@kaist.ac.kr}
}
\maketitle
\begin{abstract}
Event cameras sense intensity changes 
and have many advantages over conventional cameras. 
To take advantage of event cameras, 
some methods have been proposed to reconstruct intensity images from event streams. However, 
the outputs are still in low resolution (LR), noisy, and unrealistic. The low-quality outputs stem broader applications of event cameras, where high spatial resolution (HR) is needed as well as high temporal resolution, dynamic range, and no motion blur. 
We consider the problem of reconstructing and super-resolving intensity images from LR events, when no ground truth (GT) HR images and down-sampling kernels are available. 
To tackle the challenges, we propose a novel end-to-end pipeline that reconstructs LR images from event streams, enhances the image qualities, and upsamples the enhanced images, called EventSR. For the absence of real GT images, our method is primarily unsupervised, deploying adversarial learning.
To train EventSR, we create an open dataset including both real-world and simulated scenes. 
The use of both datasets boosts up the network performance, and the network architectures and various loss functions in each phase help improve the image qualities. The whole pipeline is trained in three phases. While each phase is mainly for one of the three tasks, the networks in earlier phases are fine-tuned by respective loss functions in an end-to-end manner.  
Experimental results show that EventSR reconstructs high-quality SR images from events 
for both simulated and real-world data. A video of the experiments is available at \url{ https://youtu.be/OShS_MwHecs}. 
\end{abstract}

\begin{figure}[t!]
\begin{center}
\vspace{-2pt}
    \includegraphics[width=\columnwidth, height=4.6cm]{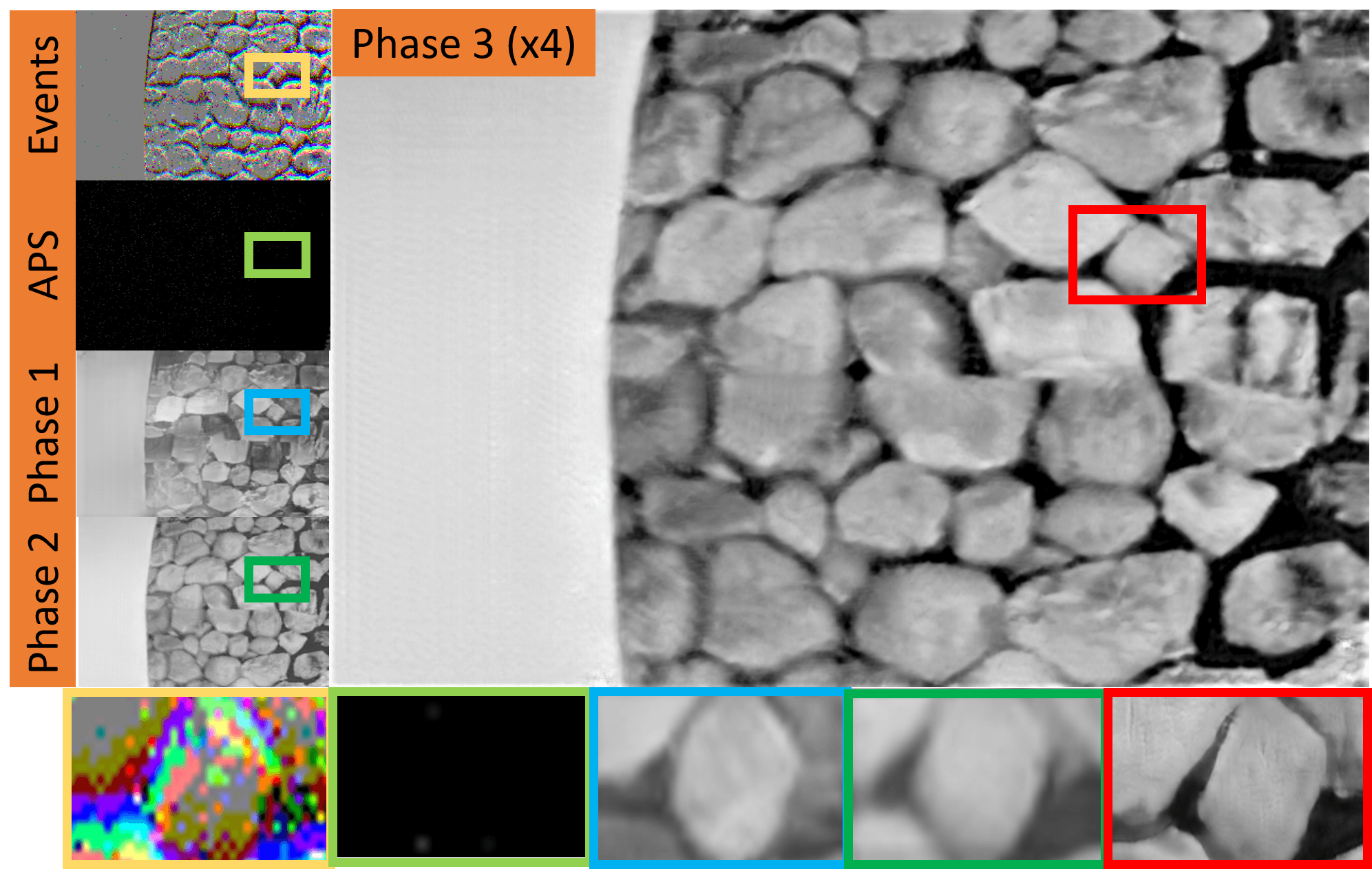}
\vspace{-15pt}
\captionsetup{font=small}
\caption{Reconstructing realistic HDR SR intensity image from pure events. EventSR reconstructs LR HDR intensity image,  restores realistic LR image and finally generates SR image (with scale factor of $\times4$) from events in phase 1,2 and 3, respectively. 
} 
\label{fig:hdr_impressive}
\end{center}
\vspace{-28pt}
\end{figure}

\vspace{-10pt}
\section{Introduction}
\label{Sec_intro}
Event cameras are bio-inspired sensors that
sense the changes of intensity at the time they occur and produce asynchronous event streams \cite{lichtsteiner2008128,wang2019event,kim2014simultaneous}, while  conventional cameras capture intensity changes at a fixed frame rate.
This distinctive feature has sparked a series of methods developed specific for event cameras~\cite{scheerlinck2018continuous}, and only recently, generic learning algorithms were successfully applied to event-based problems~\cite{wang2019event, zhu2019unsupervised,wang2019ev,rebecq2019events,cannici2019asynchronous}. 

Event cameras (\eg, DAVIS 240) convey clear advantages such as very high dynamic range (HDR) ($140dB$) \cite{lichtsteiner2008128}, no motion blur and high temporal resolution ($1\mu$s), and it has been shown that an event camera alone is sufficient to perform high-level tasks such as object detection \cite{mitrokhin2018event}, tracking \cite{gehrig2018asynchronous}, and SLAM \cite{kim2016real}. In addition, as its potential, event streams might contain complete visual information for reconstructing high quality intensity images and videos with HDR and no motion blur. 
However, state-of-the-arts (SOTA) \cite{wang2019event,rebecq2019events, munda2018real, bardow2016simultaneous} for intensity image reconstruction 
suffer due to accumulated noise and blur (out of focus) in stacked events and low resolution (LR) of event cameras. The active pixel sensor (APS) images are with low dynamic range, LR and blur. The reconstructed images thus typically are in LR and with artifacts. 
Although \cite{kim2016real,reinbacher2017real} focused HR for event cameras, namely spherical HR image mosaicing and HR panorama of events, respectively, they did not consider image-plane HR intensity image reconstruction and its perceptual realisticity.

In this work, we strive to answer the question, `\textit{is it possible to directly super-resolve LR event streams to reconstruct image-plane high quality high resolution (HR) intensity images}?' 
The challenges aforementioned render the reconstruction of HR intensity images ill-posed.
The problem of reconstructing, restoring (\eg denoising/deblurring), and super-resolving intensity images from pure event streams has not been convincingly excavated and substantiated.
We delve into the problem of reconstructing high-quality SR intensity images with HDR and no motion blur. 

For conventional camera images, deep learning (DL) based methods have achieved significant performance gains on single image super-resolution (SISR) using LR and HR image pairs \cite{sajjadi2017enhancenet,ledig2017photo,wang2018esrgan}. Most of the works assume that the downsampling methods are available and LR images are pristine. When it comes to event cameras, either stacked events or APS images are noisy and blurred, and GT HR images are unavailable, let alone the degradation models. It is less clear if such DL methods work for event cameras. 

Inspired by the development of DL on image translation \cite{zhu2017unpaired, wang2020deceiving}, denoising/debluring \cite{zhang2010denoising,kupyn2018deblurgan}, and SISR \cite{yuan2018unsupervised, zhao2018unsupervised}, and some recent successes in DL on event camera data \cite{zhu2019unsupervised, wang2019event}, we probe unsupervised adversarial learning to the problem of reconstructing HR intensity images from LR event streams.  
The results obtained demonstrate the efficacy of our method. To the best of our knowledge, this is the \emph{first} work for   recontructing HR intensity images by super-resolving LR event streams. 
%
The proposed pipeline consists of three major tasks. First, 1) we reconstruct LR images from LR event streams. However, these reconstructed images are usually noisy, blurred and unrealistic. 2) So, we then restore (deblur/denoise) 
realistic LR intensity images from events. 3) Finally, we super-resolve the restored LR images to SR 
images from events as shown in Fig.~\ref{fig:hdr_impressive}. 
Our framework is an end-to-end learning approach and, for more efficient training, we propose phase-to-phase network training strategy. The losses of later phases are back-propagated to the networks of earlier phases. The various loss functions and detailed network architectures are also important to best qualities.  
We build an open dataset containing $110K$ images for event to SR image reconstruction, using an event camera simulator \cite{rebecq2018esim}, event camera dataset \cite{mueggler2017event}, and also RGB SR dataset \cite{zeyde2010single,timofte2018ntire}. The conjunctive and alternative use of both real-world and simulated data for EventSR effectively boosts up the network performance. 
Experimental results using both the simulated dataset \cite{wang2019event} and real-world dataset \cite{mueggler2017event} show that EventSR achieves significantly better results than the SOTAs \cite{wang2019event, bardow2016simultaneous, munda2018real}. 
In summary, our contributions are: 1) the \emph{first} pipeline of reconstructing image-plane HR intensity images from LR events considering image restoration, 2) an \emph{open} dataset to train EventSR for event-based super-resolution and the \emph{skills} of using it for high performance training, 3) the proposed detail architecture, loss functions and end-to-end learning strategy, and 4) \emph{better} results than the SOTA works for image reconstruction. Our dataset is open at \url{https://github.com/wl082013/ESIM_dataset}.  

\vspace{-2pt}
\section{Related Works}
\noindent\textbf{Events to intensity image reconstruction}
The first attempt that reconstructed intensity images from rotating event cameras was done by  \cite {cook2011interacting} and \cite{kim2014simultaneous}  with rotation of visual representations.
Later on,  \cite{kim2016real} further delved into reconstructing HR masaic images based on spherical 3D scenes and estimated 
6 degree-of-freedom (6DoF) 
camera motion . Besides, Bardow \etal \cite{bardow2016simultaneous} proposed to estimate optical flow and intensity changes simultaneously via a variational energy function. Similarly, Munda \etal \cite{munda2018real} regarded image reconstruction as an energy minimization problem defined on manifolds induced by event timestamps. Compared to \cite{munda2018real}, Scheerlinck \etal \cite{scheerlinck2018continuous} proposed to filter events with a high-pass filter prior to integration. Recently, DL-based approaches brought great progress on intensity image and video reconstruction. 
Wang \etal \cite{wang2019event} proposed to use GAN \cite{goodfellow2014generative, bardow2018estimating, wang2020deceiving} to reconstruct intensity images and achieved the SOTA performance. In contrast, Rebecq \etal \cite{rebecq2019events} exploited recurrent networks to reconstruct video from events. They also used an event sensor with VGA ($640 \times 480$ pixels) resolution to reconstruct higher resolution video, however, the problem is essentially different from our work.

\noindent\textbf{Deep learning on event-based vision} 
\cite{reinbacher2017real} considered sub-pixel resolution to create a panorama for tracking with much higher spatial resolution of events, however, not of intensity image reconstruction.
Alonso \etal \cite{alonso2019EvSegNet} further used an encoder-decoder structure for event segmentation. 
In contrast, Zhu \etal \cite{zhu2019unsupervised} utilized an encoder-decoder network for optical flow, depth and ego-motion estimation via unsupervised learning. 
Besides, Cannici \etal \cite{cannici2019asynchronous} refined YOLO \cite{redmon2016you} for event-based object detection. Moreover, \cite{wang2019ev} and \cite{calabrese2019dhp19} both utilized CNNs for human pose estimation and action recognition.
Meanwhile, to analyze event alignment, Gallego \etal \cite{Gallego_2018_CVPR, gallego2019focus} proposed some loss and optimization functions,which are further applied to motion compensation \cite{stoffregen2019event}, flow estimation \cite{zhu2019unsupervised}, etc. 

\begin{figure*}[th]
    \centering
    \includegraphics[ width=1.0\textwidth]{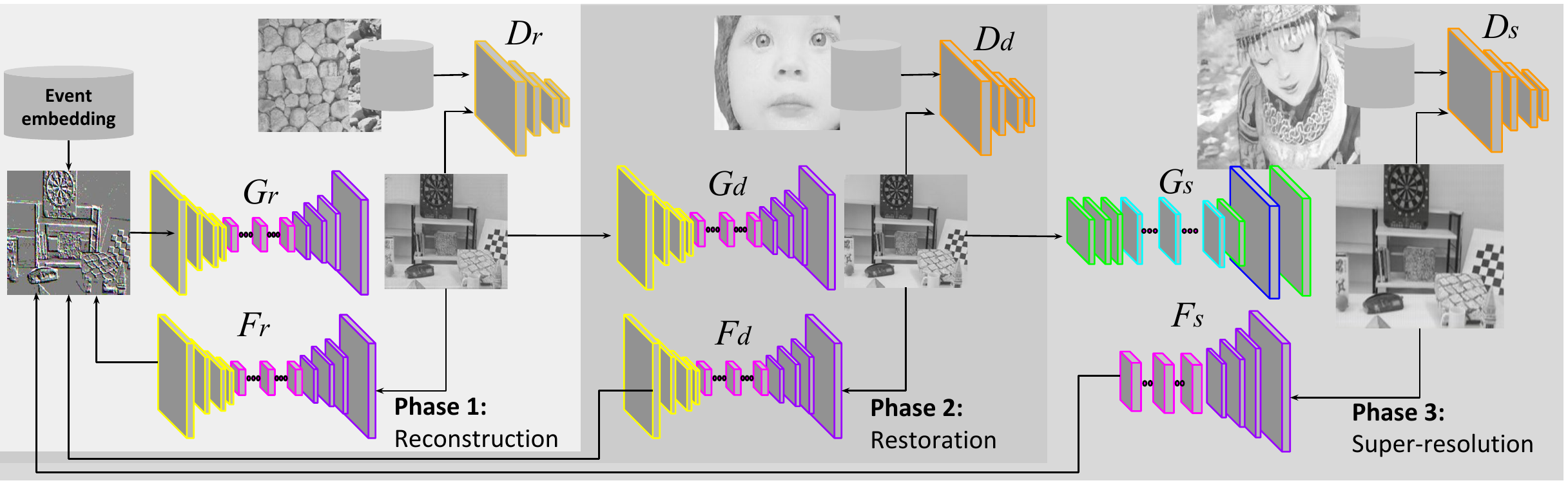}
    \vspace{-18pt}
    \captionsetup{font=small}
    \caption{
    An illustration of the proposed EventSR consisting of three phases: event to image reconstruction (Phase 1), event to image restoration (Phase 2), and event to image super-resolution (Phase 3) via unsupervised adversarial learning.
    With well designed training and test dataset, EventSR not only works well for simulated but also for real-world data with HDR effects and motion blur.
}
\label{fig:method}
    \vspace{-10pt}
\end{figure*}

\noindent\textbf{Deep learning on image restoration/enhancement} 
Image restoration addresses the problem of unsatisfactory scene representation, and the goal is to manipulate an image in such a way that it will in some sense more closely depict the scene that it represents \cite{reeves2014image} by deblurring and denoising from a degraded version. While the objective of image enhancement is to process the image (\eg contrast improvement, image sharpening, super-resolution) so that it is better suited for further processing or analysis \cite{anwar2019deep}. Recently, CNN has been broadly applied to image restoration and enhancement. The pioneering works include a multilayer perception for image denoising \cite{burger2012image} and a three-layer CNN for image SR \cite{dong2014learning}. Deconvolution was adopted to save computation cost and accelerate inference speed \cite{dong2016accelerating, shi2016real}. Very deep networks were designed to boost SR accuracy in \cite{kim2016accurate,lim2017enhanced}. Dense connections among various residual blocks were
included in \cite{zhang2018residual}. Similarly, CNN- and GAN-based methods were developed for image denoising in \cite{liu2018when,kupyn2018deblurgan, yuan2018unsupervised,zhao2018unsupervised}.

\vspace{-3pt}
\section{Proposed Methods}
\vspace{-2pt}
Our goal is to reconstruct SR images $\mathcal{I}^{SR}$ from a stream of events $\mathcal{E}$. 
To feed events to the network, we consider merging events based on the number of incoming events to embed them into images as done in \cite{wang2019event, zhu2019unsupervised}.
We then propose a novel unsupervised framework that incorporates namely, event to image reconstruction (Phase 1), event to image restoration (Phase 2), and event to image super-resolution (Phase 3) as shown in Fig.~\ref{fig:method}. We train the whole system in a sequential phase-to-phase manner, than learning all from scratch. This gradually increases the task difficulty to finally reconstruct SR images. In each phase, the networks of earlier phases are updated thus in an end-to-end manner. More details 
are given in Sec.~\ref{sec_loss}.

\subsection{Event embedding and datasets}
\label{data_sec}
\begin{figure}[t]
    \centering
    \includegraphics[width=\columnwidth]{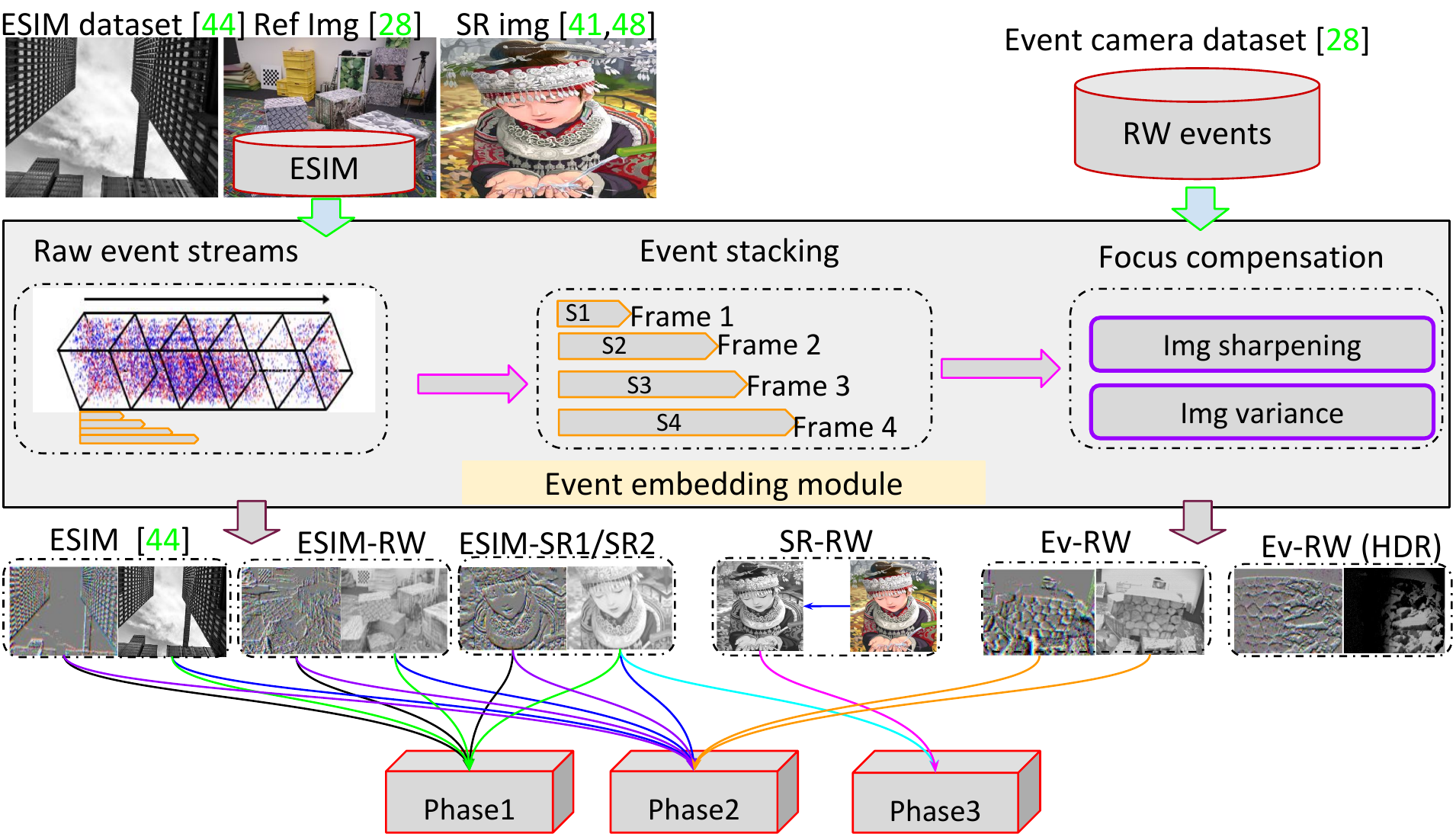}
    \vspace{-22pt}
    \captionsetup{font=small}
    \caption{An illustration of event embedding and dataset creation for  training EventSR. More details are in the main context. 
    }
    \label{fig:event_stack}
\vspace{-16pt}
\end{figure}

\noindent \textbf{Event embedding} 
To process event streams using CNNs, we need to stack events into an image or fixed tensor representation as in \cite{wang2019event, zhu2019unsupervised}. An event camera 
interprets the intensity changes as asynchronous event streams. An event \emph{e} is represented as a tuple $(u, t, p)$, where $u= (x,y)$ are pixel coordinates, $t$ is the timestamp of the event, and $p= \pm 1$ is the polarity indicating the sign of brightness change. A natural choice is to encode the events in a spatial-temporal 3D volume or voxel grid \cite{zhu2019unsupervised, wang2019event}. Here, we consider representing 3D event volume by merging events based on the number of events as shown in~Fig \ref{fig:event_stack}. We reduce event blur (out of focus) by adjusting event sharpness and also variance (contrast) as in \cite{gallego2019focus}. The first $N_e$ events are merged into frame one, and next $N_e$  are merged into frame 2, which is repeated up to frame $n$ to create one stack with $n$ frames. Thus, the stack that contains $nN_e$ events will be fed as input to EventSR. In Fig~\ref{fig:event_stack}, $S_1$, $S_2$, $S_3$ and $S_4$ are the frames containing different number of events $N_e$, $2N_e$, $3N_e$, $4N_e$, respectively. The event embedding method guarantees rich event data as inputs for EventSR and allows us to adaptively adjust $N_e$ in each frame and  $n$ in one stack.

\noindent \textbf{EventSR dataset}
One crucial contribution of this work is to build a dataset including both simulated and real-world scenes for training EventSR.
As mentioned in Sec.~\ref{Sec_intro}, real events are noisy and out of focus. Besides, real  APS  images are degraded with blur, noise and low dynamic range. Therefore, training with only real-world data is not optimal, also shown in \cite{wang2019event}, and not enough to reconstruct SR images. 
We propose a novel EventSR dataset including both real and simulated events. We utilize both data conjunctively and alternatively in each phase of training as shown in Fig.~\ref{fig:event_stack} and Table.~\ref{table:data_table}, and demonstrate that it works well. 
For the simulated data, there are three categories for different purposes. First, we use the dataset proposed by \cite{wang2019event} for comparisons in  intensity image reconstruction. Second, in order to better handle the ill-posed problem caused by real-world data \cite{wang2019event}, we utilize the reference color images from the event camera dataset \cite{mueggler2017event}. This brings a simulated dataset called ESIM-RW (around $60K$) using the event simulator(ESIM) \cite{rebecq2018esim}. The networks trained using the dataset generalises well to real event data. We also take the standard RGB SR dataset \cite{zeyde2010single, timofte2018ntire} to make ESIM-SR dataset (around $50K$). 
\emph{However, note that ESIM generates multiple synthetic events and APS images (cropped and degraded) given one HR color image, which renders the SR problem without real GT, thus making it difficult to evaluate the quality of reconstructing SR images from events.} 
\begin{table}[t]
\captionsetup{font=small}
\caption{Data source used for training EventSR. (R/S for real/synthetic, P1/P2/P3 for phase 1/2/3, Eval for numerical evaluation, Gen. for generalization to real data, \checkmark/ \xmark for yes/no, and \redcheck~ indicates very crucial for training EventSR.)}
\vspace{-20pt}
\footnotesize
\begin{center}
\begin{tabular}{p{1.75cm}|p{1.21cm}|p{0.3cm}|p{0.4cm}|p{0.3cm}|p{0.3cm}|p{0.35cm}|p{0.3cm}}
\hline
 Data name& Resolution & R/S & P1 & P2 & P3 &Eval & Gen. \\
\hline\hline
 ESIM data~\cite{wang2019event}  &256x256 & S & \checkmark & \checkmark &\checkmark& \checkmark & \xmark\\
 \hline
 ESIM-RW  &256x256 & S & \redcheck & \redcheck &\checkmark& \xmark &\checkmark\\
 \hline
 ESIM-SR1  &256x256 & S & \checkmark & \checkmark &\checkmark& \xmark & \checkmark\\
 \hline
 ESIM-SR2  &1024x1024 & S & \checkmark & \checkmark &\redcheck & \checkmark & \xmark\\
 \hline
 Ev-RW(HDR) &256x256 & R & \xmark & \redcheck & \xmark& \checkmark & \checkmark\\
 \hline
 SR-RW  &1024x1024 & R & \xmark  & \redcheck &\redcheck& \xmark & \checkmark\\
\hline
\end{tabular}
\end{center}
\vspace{-20pt}
\label{table:data_table}
\end{table}
We use ESIM to create ESIM-SR1 dataset with image size ($256 \times 256$) for training phase 1 and phase 2. 
To numerically evaluate the SR quality, we create ESIM-SR2 dataset, where we set ESIM to output `HR' APS images with larger size (\eg 1024x1024) as shown in Table.~\ref{table:data_table}, \footnote{Different from generic SR problems, these `HR' APS images are in low quality (unclear edges and corners) due to the inherent properties of event camera.  They, however, can be used to evaluate the quality of SR.} which are then downsampled (\eg bicubic) to smaller size (\eg 256x256) as LR images. However, reconstructing LR images up to the quality level of these `HR' APS images does not achieve our goal since we want to generate realistic SR images. Thus, we exploit a real-world dataset for phase 3. 
For the real-world dataset, we directly make Ev-RW dataset using the event camera dataset \cite{mueggler2017event} including general, motion blur and HDR effects. It has been shown that using real event and APS pairs for reconstructing SR images is difficult~\cite{wang2019event}. Instead, in phase 1, we use ESIM-RW dataset, which is crucial for training EventSR. In phase 2, we first refine the real APS images through phase 1 to get clean APS images (the reason is given in Eq.~(\ref{id_loss})), then use them for event to image restoration. Lastly, for phase 3, we convert the RGB SR images to grayscale as SR-RW dataset, and it turns out they are crucial for training EventSR. The trained EventSR generalizes well for both simulated and real data, and also the data with HDR effects as shown in Fig.~\ref{fig:event_stack} and Table.~\ref{table:data_table}. 

\vspace{-2pt}
\subsection{Loss functions and training strategy of EventSR}
\label{sec_loss}
\vspace{-2pt}
As shown in Fig.~\ref{fig:method}, EventSR consists of three phases: event to image reconstruction, event to image restoration and event to image super-resolution. EventSR includes three network functionals $G$, $F$, and $D$ in each phase.

\vspace{3pt}
\noindent\textbf{Event to image reconstruction (Phase 1)} In order to obtain SR images, we first reconstruct images from the event streams. Our goal is to learn a mapping $\mathcal{I}^{LR} = G_{r}(\mathcal{E})$, aided by an event feedback mapping $\mathcal{E} = F_{r}(\mathcal{I}^{LR})$, and the discriminator $D_{r}(\mathcal{I}^{LR})$. The inputs are unpaired training events $\mathcal{E}$ and the LR intensity images $\mathcal{I}^{LR}$. 

\vspace{3pt}
\noindent\textbf{Event to image restoration (Phase 2)} Since the reconstructed images are noisy, blurry, and unrealistic, we then aim to restore (denoise/ deblur) images using both events $\mathcal{E}$ and clean LR images $\mathcal{I}^{cLR}$. The goal of phase 2 is to learn a mapping $\mathcal{I}^{cLR} = G_{d}(G_{r}(\mathcal{E}))$, an event feedback mapping $\mathcal{E} = F_{d}(\mathcal{I}^{cLR})$, and the discriminator $D_{d}(\mathcal{I}^{cLR})$. The inputs are unpaired events $\mathcal{E}$ and the clean images $\mathcal{I}^{cLR}$.

\vspace{3pt}
\noindent\textbf{Event to image super-resolution (Phase 3)} We then reconstruct SR images from events, utilizing the stacked events $\mathcal{E}$ and real-world HR images $\mathcal{I}^{HR}$. The problem is to learn a mapping $\mathcal{I}^{SR} = G_{s}(G_{d}(G_{r}(\mathcal{E})))$, an event feedback mapping $\mathcal{E} = F_{d}(\mathcal{I}^{SR})$, and the discriminator $D_{d}(\mathcal{I}^{SR})$. \vspace{-5pt}
\vspace{-5pt}
\subsubsection{Loss functions for EventSR training}
\vspace{-5pt}
The loss functional for each phase is defined as a linear combination of four losses as:  
\begin{equation}
\resizebox{1.0\hsize}{!}{$\mathcal{L} =
    \mathcal{L}_{Adv}(\bar{G},D) + \lambda_1  \mathcal{L}_{Sim}(\bar{G},F) +  \lambda_2 \mathcal{L}_{Id}(\bar{G}) +   \lambda_3 \mathcal{L}_{Var}(\bar{G})$
}\label{sum}\end{equation}
where $\mathcal{L}_{Adv}$, $\mathcal{L}_{Sim}$, $\mathcal{L}_{Id}$, $\mathcal{L}_{Var}$ are the discriminator, event similarity, identity, and total variation losses, respectively. Note $D$ and $F$ are the relevant networks of each stage, and $\bar{G}$ is an  accumulated one i.e. $G_r, G_d(G_r), G_s(G_d(G_r))$, in phase 1, 2, and 3. The  loss for phase 1,2 and 3 is denoted as $\mathcal{L}^r$, $\mathcal{L}^d$ and $\mathcal{L}^s$, respectively. 

\vspace{3pt}
\noindent\textbf{Adversarial loss $\bm{\mathcal{L}_{Adv}}$}  
Given stacked events $\mathcal{E}$, the generator $\bar{G}$ learns to generate what are similar to given the dataset i.e. the reconstructed, the restored, and the super-resolved, respectively. The discriminator $D$ in this case learns to distinguish the generated images from the given target images via discriminator loss $\mathcal{L}_D$. The adversarial loss is:
\begin{equation}
\label{gan_loss}
   \mathcal{L}_{Adv}(\bar{G},D) =  -\E[log(1- D(\bar{G}(\mathcal{E})))].
\end{equation}
We observe standard GAN training is difficult in phase 3. To stabilize the training and make optimization easier, we use the adversarial loss based on the Relativistic GAN \cite{jolicoeur2018relativistic}. 

\vspace{3pt}
\noindent\textbf{Event similarity loss $\bm{\mathcal{L}_{Sim}}$} Since events are usually sparse, we found using pixel-level loss too restrictive and less effective. Here, we propose a new event similarity loss that is based on the interpolation of the pixel-level loss and the perceptual loss  based VGG19 inspired by \cite{johnson2016perceptual}. Namely, we measure the similarity loss of the reconstructed events $F(\bar{G}(\mathcal{E}))$ and the input events $\mathcal{E}$. We linearly interpolate the pixel-wise loss like $L_2$ and the perceptual loss as:
\begin{equation}
\label{sim_loss_equ}
\begin{split}
    \mathcal{L}_{Sim}(\bar{G},F) =  \E \Big[ \alpha ||F(\bar{G}(\mathcal{E}))- \mathcal{E}||_2 +\\ (1-\alpha) \frac{1}{C_iW_iH_i}||\Phi_i(F(\bar{G}(\mathcal{E}))) -\Phi_i(\mathcal{E})||_2 \Big]
\end{split}
\end{equation} 
where $\Phi_i$ is the feature map from $i$-th VGG19 layer, and $C_i$, $H_i$, and $W_i$ are the number of channel, height, and width of the feature maps, respectively.

\vspace{3pt}
\noindent\textbf{Identity loss $\bm{\mathcal{L}_{Id}}$ }
For better learning from events, and also to avoid brightness and contrast variation among different iterations, we utilize the identity loss $\mathcal{L}_{Id}$. \emph{Besides, since Ev-RW APS images are noisy, we use $\mathcal{L}_{Id}$ to optimize $G_{r}$ as a denoiser using clean synthetic APS images. When $G_{r}$ is trained, the Ev-RW APS images are fed to the denoiser to get clean real-world images $\mathcal{I}^{cLR}$ to train $G_{d}$ in phase 2}.  
\begin{equation}
\label{id_loss}
    \mathcal{L}_{Id}(\bar{G}) = \E[||\bar{G}(\mathcal{I}) - \mathcal{I}||_2]
\end{equation} 
where $\mathcal{I}$ and $\bar{G}$ are the target image and the generator in each phase. Since there is the upsampling operation in $G_s$ of phase 3, we propose to use the downsampled HR images as input to $G_s$. The identity loss helps preserve the shading and texture composition between the $\bar{G}(\mathcal{I})$ and $\mathcal{I}$. 

\vspace{3pt}
\noindent\textbf{Total variation loss $\bm{\mathcal{L}_{Var}}$} Since stack events are sparse, the generated images are spatially not smooth. To impose the spatial smoothness of the generated images, we add a total variation loss:
\begin{equation}
\label{tv_loss}
    \mathcal{L}_{Var}(\bar{{G}}) = \E[||\nabla_h \bar{G}(\mathcal{E}) + \nabla_w \bar{G}(\mathcal{E})||_2],
\end{equation} 
where $\nabla_h$ and $\nabla_w$ are the gradients of $\bar{G}$. 

\subsubsection{Learning strategy and network structure}
\noindent \textbf{End-to-end learning }
We have described the pipeline of reconstructing, restoring, and attaining SR images from events. We then explore how to unify three phases and train EventSR in an end-to-end manner. Under the unified learning, the second phase becomes auxiliary to the first and the third stage auxiliary to the second and the first. The total loss is:
\begin{equation}
    \mathcal{L}^{total} = \mathcal{L}^{r} + \mathcal{L}^{d} + \mathcal{L}^{s}
\end{equation}

\noindent \textbf{Phase-to-phase learning}
Rather than learning all network parameters from scratch all together, to facilitate the training, we propose a learning strategy called phase-to-phase learning where we start with an easy task and then gradually increase the task difficulty. Specifically, we first start with $G_r$ with $D_r, F_r$. We then strengthen the task difficulty by fusing $G_r$ and $G_d$. We train $G_d$ and $D_d, F_d$ from scratch, meanwhile, fine-tuning ${G}_r$. Note each loss term has $\bar{G}$ which is the cascaded reconstruction function i.e. $G_d(G_r)$ in the phase 2. The loss gradients back-propagated to $G_r$, and $D_r, F_r$ are also updated respectively. We lastly fuse $G_s$ with both $G_r$ and $G_d$ from events. We train the $G_s, D_s, F_s$ from scratch, while fine-tuning both $G_r, G_d$ simultaneously. The generation function $\bar{G} = G_s(G_d(G_r))$. 

\vspace{3pt}
\noindent \textbf{Network Architecture}
As shown in Fig.~\ref{fig:method}, EventSR includes three generators, $G_r$, $G_d$ and $G_s$, and three discriminators, ${D}_r$ and ${D}_d$ and ${D}_s$. For convenience and efficiency, we design $G_r$, $G_d$ to share the same network structure. For ${G}_s$, we adopt the SOTA SR networks \cite{wang2018esrgan,ledig2017photo}. We also set ${D}_r$, ${D}_d$, and ${D}_s$ to share the same network architecture. 
To better utilize the rich information in events, we also design an event feedback module including ${F}_r,{F}_d$, and ${F}_s$, sharing the same network structures based on ResNet blocks. However, for ${F}_s$, it has down-sampling operation, so we set the stride with 2 instead. 
Through the event feedback module, the generators learn to fully utilize the rich information from events to reconstruct, restore, and super-resolve images from events. 

\section{Experiments and Evaluation}
\vspace{-1pt}
\noindent\textbf{Implementation and training details} To facilitate the efficient training of our network, we utilize the proposed phase-to-phase training strategy to achieve the goal of end-to-end learning. In phase 1, we train ${G}_{r}$ and  ${D}_r$ with feedback network ${F}_{r}$. We set $\alpha =0.6$ in Eq.~\ref{sim_loss_equ} and $\lambda_1= 10$, $\lambda_2=5$ and $\lambda_3=0.5$ in Eq.~\ref{sum}. We then train ${G}_d({G}_r(\mathcal{E})$) and ${D}_d$ from scratch with ${F}_d$ in phase 2.  We set the $\lambda_1= 10$, $\lambda_2=5$ and $\lambda_3=2$ in Eq.~\ref{sum}. In phase 3, we train  and  ${G}_r ({G}_d({G}_s(\mathcal{E})))$ from scratch with ${F}_s$. The parameters in this phase are set with $\lambda_1 =10$, $\lambda_2=5$ and $\lambda_3=3$ in Eq.~\ref{sum}.  We initialize the network with dynamic learning rate. we set the batch size of 1 for single GPU, and augment the training data by random rotation and horizontal flipping. We use Adam solver \cite{kingma2014adam} with $\beta_1 = 0.9$ and $\beta_2=0.999$ to optimize our framework. $3$ stacks ($N_e=10K$ events per stack) are used to get an event image. 
We assess the quality of each phase outputs using the SSIM, FSIM, and PSNR. To compare with SOTA works~\cite{rebecq2019events, munda2018real, bardow2016simultaneous}, we also use LPIPS \cite{zhang2018unreasonable} for measuring the image quality. For all datasets, in order to measure the similarity, each APS image is matched with corresponding reconstructed images with the closest timestamp. We mainly focus on scaling factor of $\times4$ since it is more challenging and meaningful as studied in SOTA SR works \cite{wang2018esrgan, wang2018esrgan}.

\begin{figure*}[t!]
\begin{center}
\vspace{-2pt}
\renewcommand{\tabcolsep}{1pt}
\begin{tabular}{@{}ccccccc@{}}
\small{APS}& \small{Stacked events} &  \small{E2VID \cite{rebecq2019events}}& \small{Wang \cite{wang2019event}}& \small{Phase 1 Rec.} &   \small{Phase 2 Rest.} &  \small{Phase 3 SR(x4)}  \\ 

    \includegraphics[width=24mm,height=17mm]{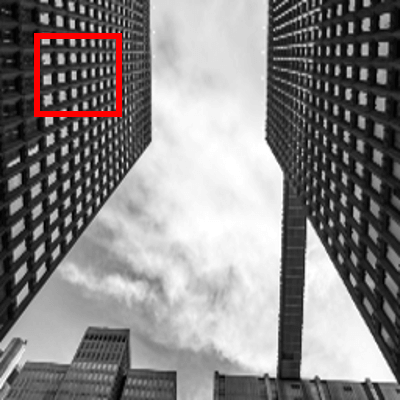}& 
    \includegraphics[width=24mm,height=17mm]{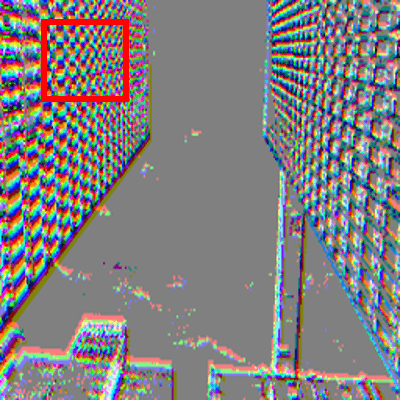}&
    \includegraphics[width=24mm,height=17mm]{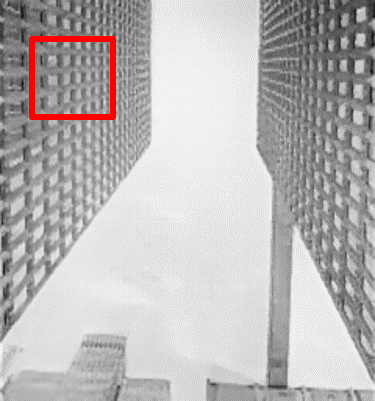}&
    \includegraphics[width=24mm,height=17mm]{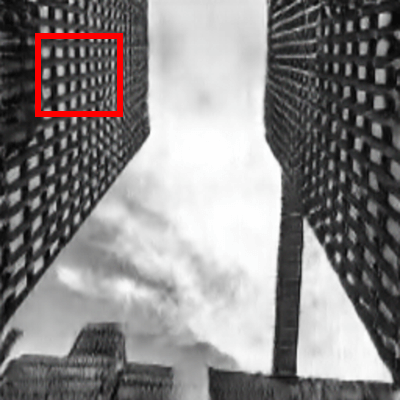}&
    \includegraphics[width=24mm,height=17mm]{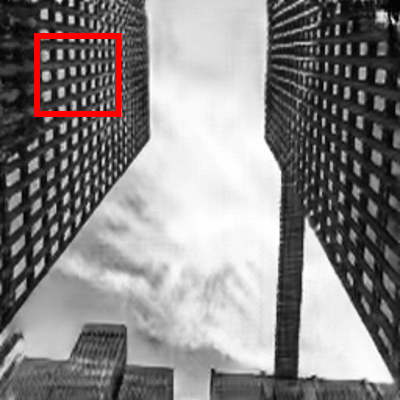}&
   
    \includegraphics[width=24mm,height=17mm]{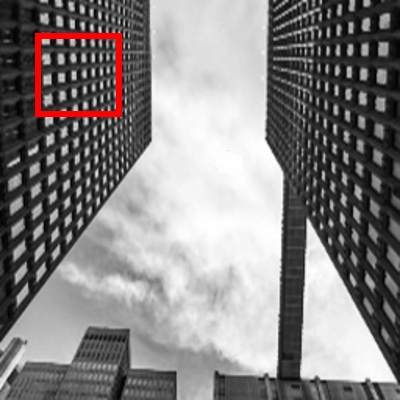}&
    \includegraphics[width=24mm,height=17mm]{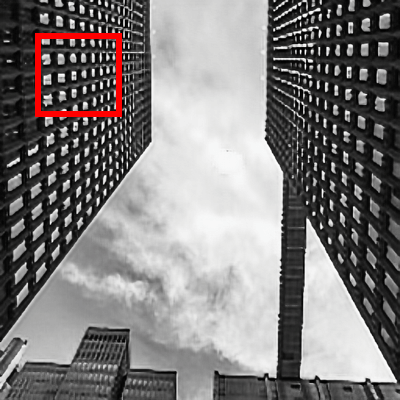}\\

    \includegraphics[width=24mm,height=13mm]{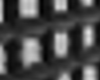}&
    \includegraphics[width=24mm,height=13mm]{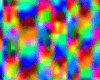}&
    \includegraphics[width=24mm,height=13mm]{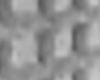}&
    \includegraphics[width=24mm,height=13mm]{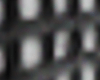}&
    \includegraphics[width=24mm,height=13mm]{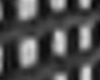}&
    \includegraphics[width=24mm,height=13mm]{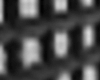}&
    \includegraphics[width=24mm,height=13mm]{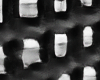}\\

\end{tabular}
\vspace{-12pt}
\captionsetup{font=small}
\caption{Visual comparison on ESIM dataset \cite{wang2019event}. The first row shows our results and the second row shows the cropped patches. EventSR achieves similar performance regarding phase 1 and better results in phase 2. 
} 
\label{fig:compare_wang_esim}
\end{center}
\vspace{-22pt}
\end{figure*}

\begin{figure}[t]
\begin{center}
\renewcommand{\tabcolsep}{1pt}
\begin{tabular}{@{}cccccc@{}}
 \small{APS}& \small{events} & \small{Phase 1} & \small{Phase 2} &  \small{Phase 3(x4)}  \\ 
    \includegraphics[width=16mm,height=17mm]{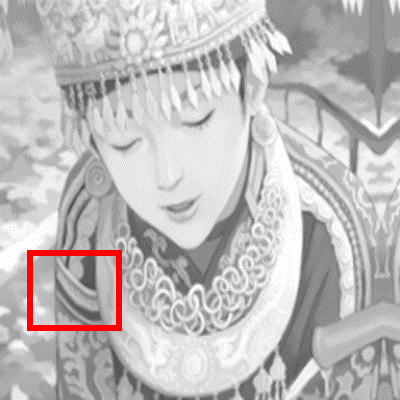}&
    \includegraphics[width=16mm,height=17mm]{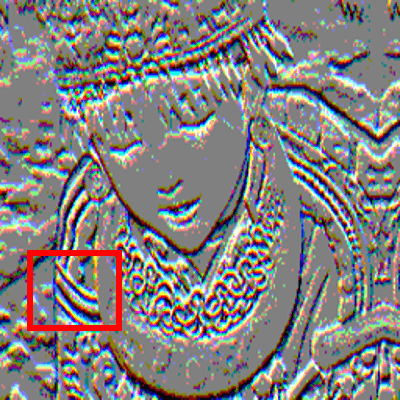} &
    \includegraphics[width=16mm,height=17mm]{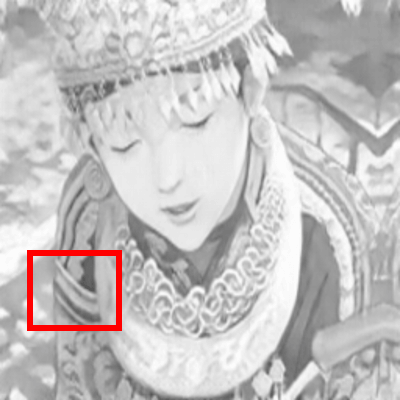} & 
    \includegraphics[width=16mm,height=17mm]{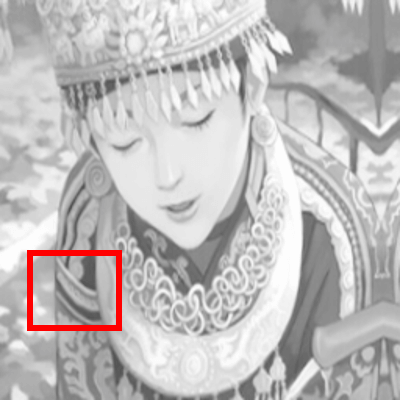}& 
    \includegraphics[width=16mm,height=17mm]{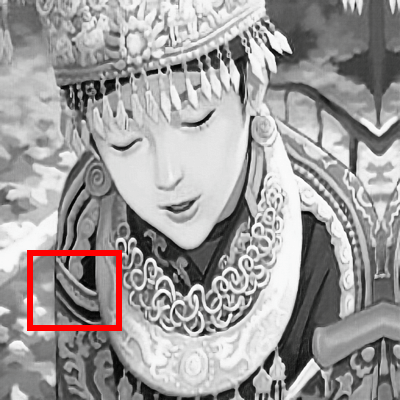}\\
    \includegraphics[width=16mm,height=13mm]{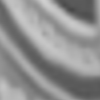} &
    \includegraphics[width=16mm,height=13mm]{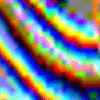} &
    \includegraphics[width=16mm,height=13mm]{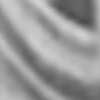} & 
    \includegraphics[width=16mm,height=13mm]{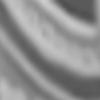}& 
    \includegraphics[width=16mm,height=13mm]{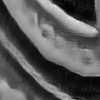}\\

\end{tabular}
\vspace{-12pt}
\captionsetup{font=small}
\caption{Visual results on our open ESIM-SR dataset. First row shows our results and the second row shows the cropped patches. 
} 
\label{fig:exp_sr}
\end{center}
\vspace{-15pt}
\end{figure}
\begin{figure}[t!]
\begin{center}
\renewcommand{\tabcolsep}{1pt}
\begin{tabular}{@{}ccccc@{}}

    \includegraphics[width=16mm,height=17mm]{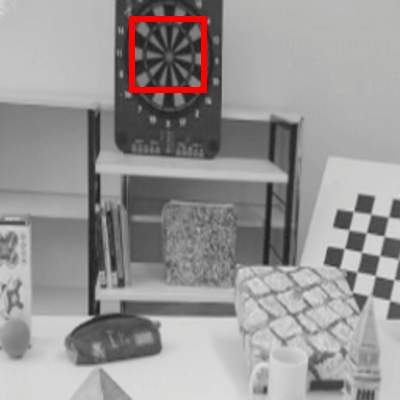}&
    \includegraphics[width=16mm,height=17mm]{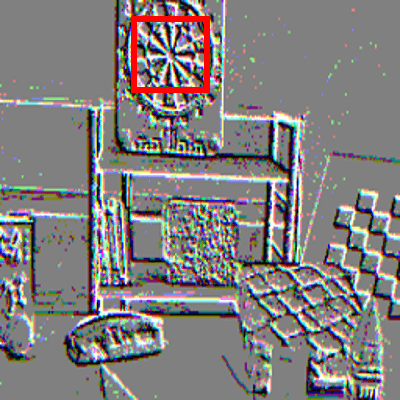}&
    \includegraphics[width=16mm,height=17mm]{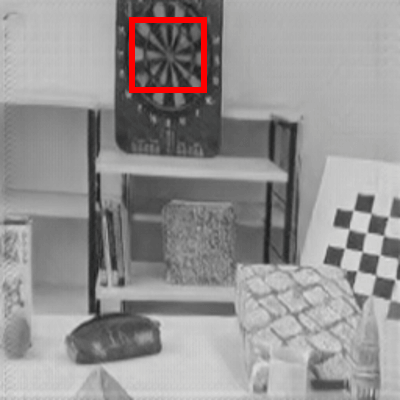}&
    \includegraphics[width=16mm,height=17mm]{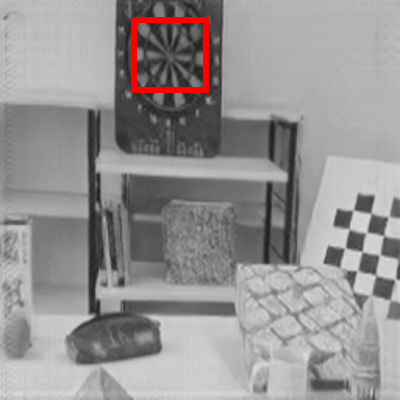}&
    \includegraphics[width=16mm,height=17mm]{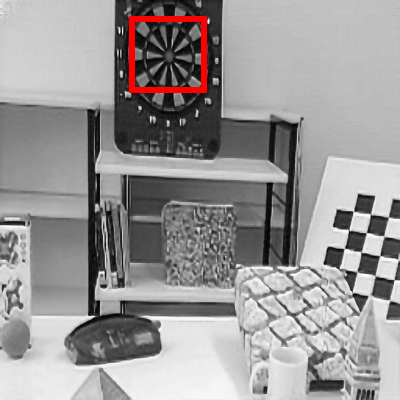}\\
    
    \includegraphics[width=16mm,height=13mm]{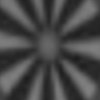}&
    \includegraphics[width=16mm,height=13mm]{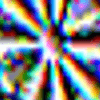}&
    \includegraphics[width=16mm,height=13mm]{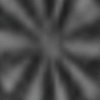}&
    \includegraphics[width=16mm,height=13mm]{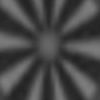}&
    \includegraphics[width=16mm,height=13mm]{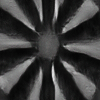}\\

\end{tabular}
\vspace{-10pt}
\captionsetup{font=small}
\caption{Results on our ESIM-RW dataset. EventSR recovers significant visual structure from events. With ESIM-RW dataset for training EventSR, it also works well on real-world events. 
} 
\label{fig:rw_esim}
\end{center}
\vspace{-15pt}
\end{figure}

\subsection{Evaluation on simulated datasets}
\label{sec:exp_sim}
\vspace{-1pt}
We first compared with \cite{wang2019event, rebecq2019events}, which is supervised-learning-based, using dataset proposed in \cite{wang2019event}. 
Figure~\ref{fig:compare_wang_esim} shows qualitative results on event to image reconstruction (phase 1), restoration (phase 2), and SR with scaling factor of 4 (phase 3). It is shown that EventSR is able to recover the lattices and alleviates the blurry artifacts, which can be visually verified in the cropped patches (second row). Besides, the generated LR images in phase 1 and 2 are close to APS image.  Table.~\ref{table:compare_esim} shows the quantitative evaluation of phase 1 and 2 results and comparison with \cite{wang2019event, rebecq2019events}. 
It turns out that our phase 1 (Ours-Rec ($n=3$)) is comparable to  \cite{wang2019event, rebecq2019events} regarding image reconstruction. Since the stacked event images are noisy and out of focus, the reconstructed images are also noisy, blurred and unrealistic. However, our phase 2 successfully handles these problems and 
achieves much better results than \cite{wang2019event, rebecq2019events} and phase 1. 

\begin{table}[t!]
\vspace{-3pt}
\captionsetup{font=small}
\caption{Quantitative comparison of phase 1 and 2 with \cite{wang2019event, rebecq2019events} (supervised) based on dataset \cite{wang2019event}. Our phase 1 achieves comparable results with \cite{wang2019event, rebecq2019events} and phase 2 achieves much better results. }
\vspace{-19pt}
\small
\begin{center}
\begin{tabular}{c|c|c|c}
\hline
 & PSNR ($\uparrow$) & FSIM ($\uparrow$) & SSIM ($\uparrow$) \\
\hline\hline
E2VID \cite{rebecq2019events} & 22.74$\pm$1.96 &  0.84$\pm$0.06  & 0.75$\pm$0.10 \\
Wang \cite{wang2019event}($n=1$) & 20.51$\pm$2.86 &  0.81$\pm$0.09  & 0.67$\pm$0.20 \\
Wang \cite{wang2019event}($n=3$)  & 24.87$\pm$3.15 &  0.87$\pm$0.06  & 0.79$\pm$0.12 \\ \hline
\rowcolor[gray]{.8} Ours-Rec ($n=3$)  & 23.26$\pm$3.60 & 0.85$\pm$0.09  & 0.78$\pm$0.24\\
\rowcolor[gray]{.8} Ours-Rest ($n=3$)  & \textbf{26.75}$\pm$2.85 &  \textbf{0.89}$\pm$0.05  & \textbf{0.81}$\pm$0.23 \\
\hline
\end{tabular}
\end{center}
\vspace{-25pt}
\label{table:compare_esim}
\end{table}

\noindent\textbf{Evaluation on ESIM-SR dataset}
We also validate EventSR on our ESIM-SR dataset. Figure~\ref{fig:exp_sr} shows the qualitative results on ESIM-SR1 dataset. Our method can recover very complex objects such as human face. We can see EventSR could utilize the high-frequency information (\eg edge/corner) in the events to reconstruct SR images better than APS images (second row). As mentioned in Sec.~\ref{data_sec}, there are no GT images for ESIM-SR1, thus making quantitative evaluation of SR images difficult. 
 However, we use the ESIM to output `HR' APS images ($1024 \times 1024$) and then downsample (\eg bicubic) them to LR images with scale factor of 4 (ESIM-SR2 dataset). So, we could quantitatively evaluate the quality of EventSR on SR as shown Table.~\ref{table:compare_sr}. Although the `LR' images do not really differ from `HR' images (high PSNR and SSIM values), our method outperforms the BI method, showing better performance.

\begin{table}[t]
\captionsetup{font=small}
\caption{Quantitative evaluation of phase 3 on our ESIM-RW dataset with BI degradation model.}
\vspace{-19pt}
\small
\begin{center}
\begin{tabular}{c|c|c}
\hline
 & PSNR ($\uparrow$) & SSIM ($\uparrow$) \\
\hline\hline
 Bicubic  & 44.27$\pm$2.56 & 0.98$\pm$0.19\\
 \hline
\rowcolor[gray]{.8} Ours-Phase.3 SR x4 ($n=3$)  & \textbf{47.68}$\pm$2.17 & \textbf{0.99}$\pm$0.12 \\
\hline
\end{tabular}
\end{center}
\vspace{-15pt}
\label{table:compare_sr}
\end{table}
\begin{table}[t!]
\captionsetup{font=small}
\caption{Quantitative comparison of phase 1 (Rec.) of EventSR with state-of-the-art works based on Ev-RW dataset \cite{mueggler2017event}.  }
\vspace{-19pt}
\small
\begin{center}
\begin{tabular}{c|c|c|c}
\hline
 & LPIPS ($\downarrow$) & FSIM ($\uparrow$) & SSIM ($\uparrow$) \\
\hline\hline
HF \cite{scheerlinck2018continuous}  &0.53& --& 0.42 \\ \hline
MR \cite{munda2018real}  &0.55& -- & 0.46 \\ \hline
E2VID \cite{rebecq2019events}  &0.42 &  --  & 0.56 \\ \hline
Wang \cite{wang2019event}($n=3$)  &-- &  0.85$\pm$0.05  & 0.73$\pm$0.16 \\ \hline
 \rowcolor[gray]{.8}  Ours-Rec ($n=3$)  & {0.35} & {0.86}$\pm$0.07  & {0.75}$\pm$0.20\\
 \rowcolor[gray]{.8}  Ours-Rest ($n=3$)  & \textbf{0.32} & \textbf{0.88}$\pm$0.09 & \textbf{0.78}$\pm$0.18\\
\hline
\end{tabular}
\end{center}
\vspace{-22pt}
\label{table:rw_comp}
\end{table}

\noindent \textbf{Results on ESIM-RW dataset} We also evaluate the performance of EventSR on our ESIM-RW dataset as mentioned in Sec.~\ref{data_sec}. This novel dataset is made using the reference color images from event camera dataset \cite{mueggler2017event}, aiming to enhance the performance of EventSR on real-world data. We train $G_r$ of phase 1 using this dataset, and surprisingly $G_r$ performs well not only on ESIM-RW events but also on real world events. Figure~\ref{fig:rw_esim} shows the experimental results on our ESIM-RW dataset. EventSR can recover the correct lines and textures from events, which can be visually verified in the cropped patches in the second row.  

\begin{figure*}[t!]
\begin{center}
\renewcommand{\tabcolsep}{1pt}
\begin{tabular}{@{}ccccccc@{}}
     \small{APS}& \small{Stacked events} & \small{E2VID \cite{rebecq2019events}}& \small{Wang \cite{wang2019event}}& \small{Phase 1 Rec.} &   \small{Phase 2 Rest.} &  \small{Phase 3 SR(x4)}  \\ 

    \includegraphics[width=24mm,height=16mm]{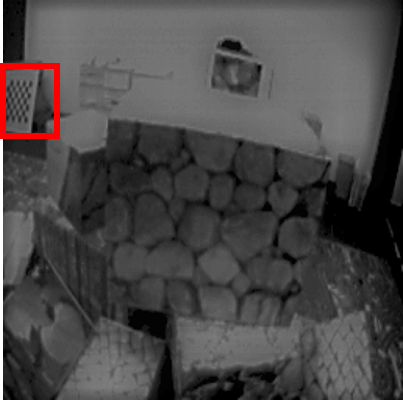} &
    \includegraphics[width=24mm,height=16mm]{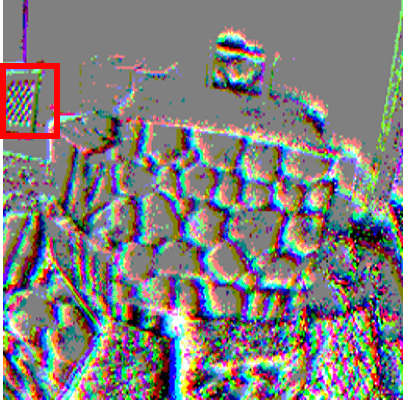}&
    \includegraphics[width=24mm,height=16mm]{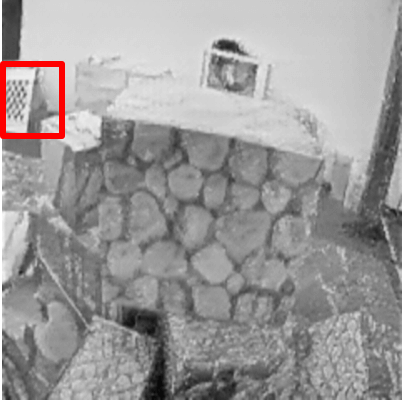}&
    \includegraphics[width=24mm,height=16mm]{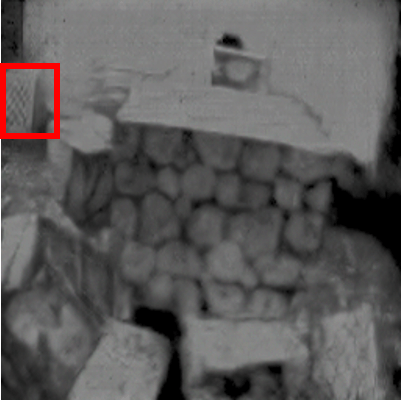}&
    \includegraphics[width=24mm,height=16mm]{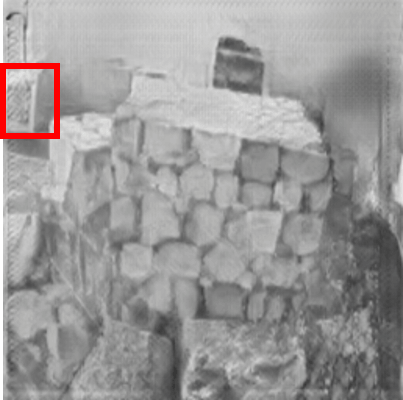}&
    \includegraphics[width=24mm,height=16mm]{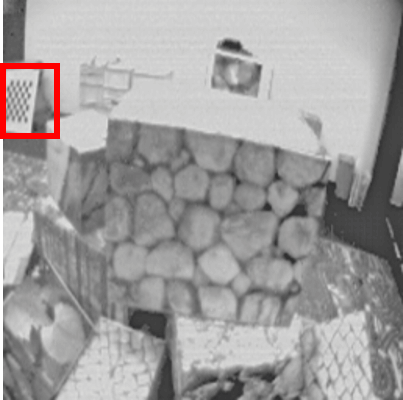}&
    \includegraphics[width=24mm,height=16mm]{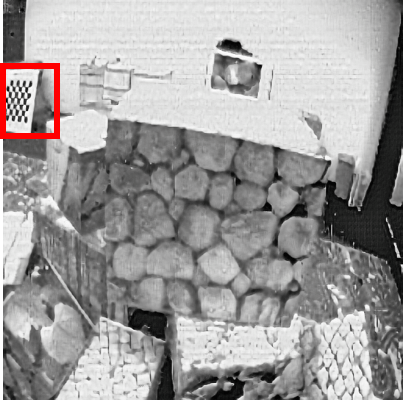}\\
    
    \includegraphics[width=24mm,height=12mm]{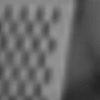}&
    \includegraphics[width=24mm,height=12mm]{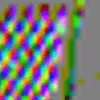}&
    \includegraphics[width=24mm,height=12mm]{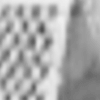}&
    \includegraphics[width=24mm,height=12mm]{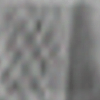}&
    \includegraphics[width=24mm,height=12mm]{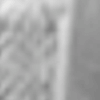}&
    \includegraphics[width=24mm,height=12mm]{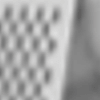}&
    \includegraphics[width=24mm,height=12mm]{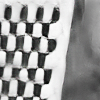} \\

    \includegraphics[width=24mm,height=16mm]{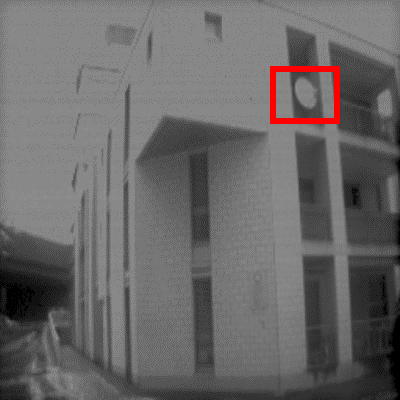}&
    \includegraphics[width=24mm,height=16mm]{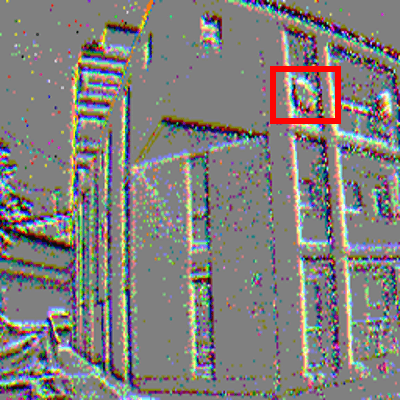}&
    \includegraphics[width=24mm,height=16mm]{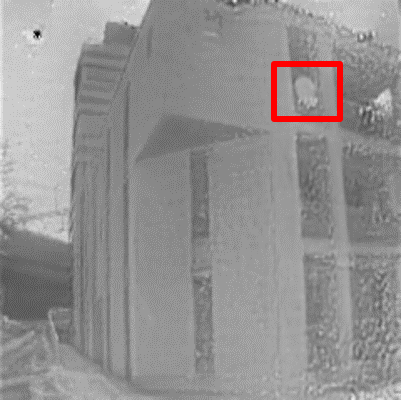}&
    \includegraphics[width=24mm,height=16mm]{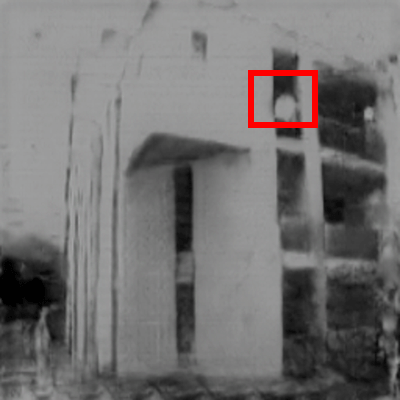}&
    \includegraphics[width=24mm,height=16mm]{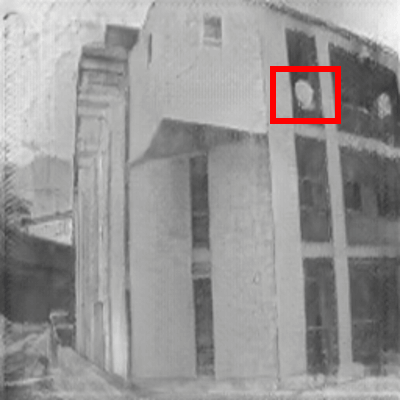}&
    \includegraphics[width=24mm,height=16mm]{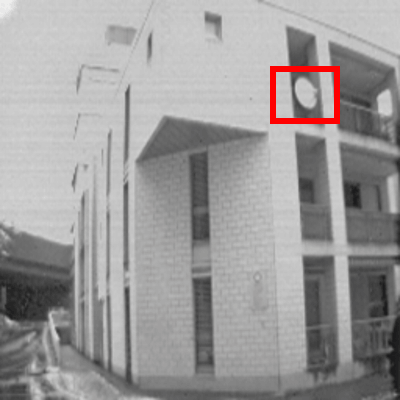}&
    \includegraphics[width=24mm,height=16mm]{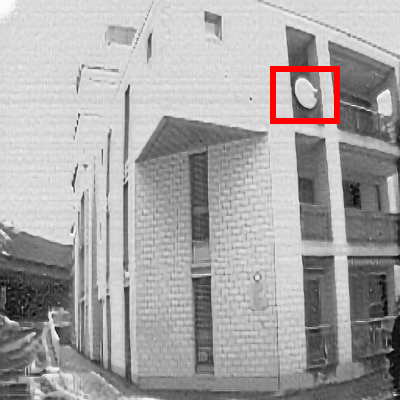}\\
    
    \includegraphics[width=24mm,height=12mm]{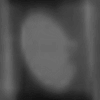}&
    \includegraphics[width=24mm,height=12mm]{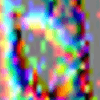}&
    \includegraphics[width=24mm,height=12mm]{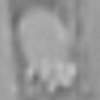}&
    \includegraphics[width=24mm,height=12mm]{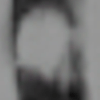}&
    \includegraphics[width=24mm,height=12mm]{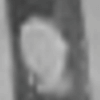}&
    \includegraphics[width=24mm,height=12mm]{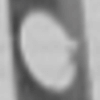}&
    \includegraphics[width=24mm,height=12mm]{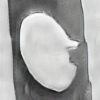}\\
\end{tabular}
\vspace{-10pt}
\captionsetup{font=small}
\caption{Visual comparison on Ev-RW dataset \cite{mueggler2017event}. With phase 1 trained using ESIM-RW, our method is capable of reconstruct the visual features like edge and corner, etc, and achieves better performance. 
} 
\label{fig:rw_exp}
\end{center}
\vspace{-16pt}
\end{figure*}

 \begin{figure*}[t!]
\begin{center}
\renewcommand{\tabcolsep}{1pt}
\begin{tabular}{@{}ccccc@{}}
 \small{APS}& \small{Stacked events} & \small{Phase 1 Rec.} & \small{Phase 2 Rest.} &  \small{Phase 3 SR(x4)}  \\ 

    \includegraphics[width=34mm,height=16mm]{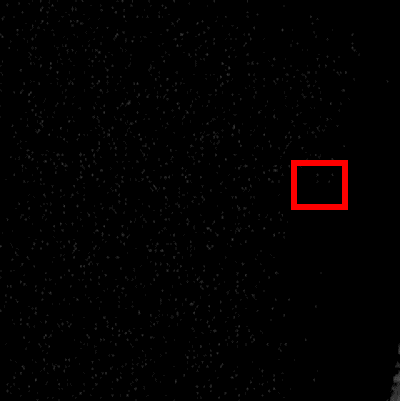}&
    \includegraphics[width=34mm,height=16mm]{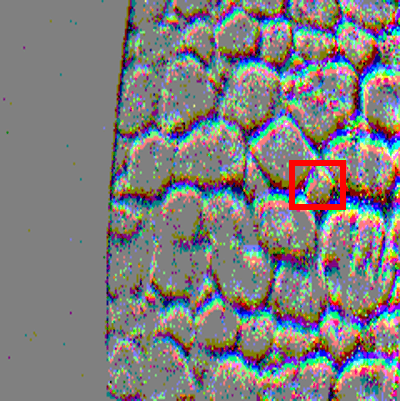}&
    \includegraphics[width=34mm,height=16mm]{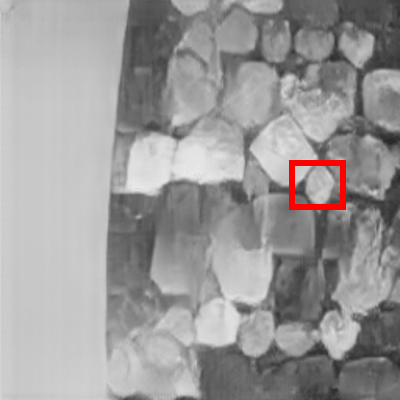}&
    \includegraphics[width=34mm,height=16mm]{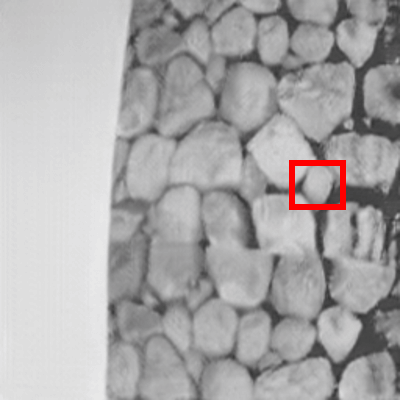}&
    \includegraphics[width=34mm,height=16mm]{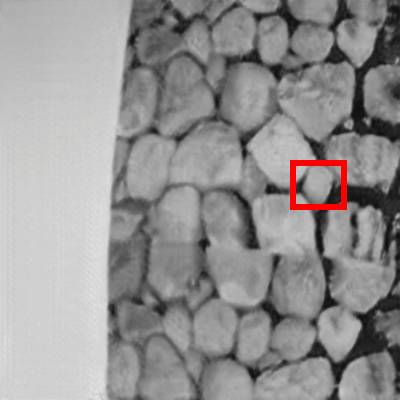}\\
    
    \includegraphics[width=34mm,height=14mm]{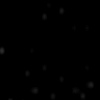}&
    \includegraphics[width=34mm,height=14mm]{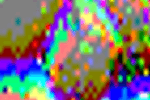}&
    \includegraphics[width=34mm,height=14mm]{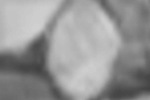}&
    \includegraphics[width=34mm,height=14mm]{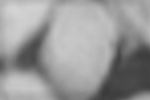}&
    \includegraphics[width=34mm,height=14mm]{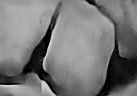}\\
 
    \includegraphics[width=34mm,height=16mm]{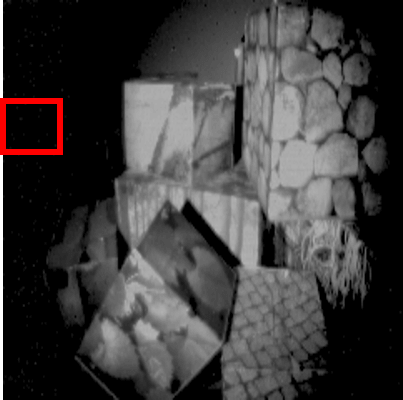}&
    \includegraphics[width=34mm,height=16mm]{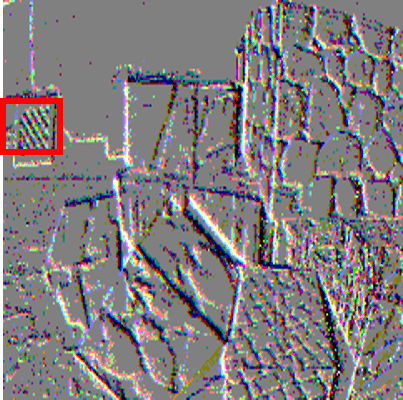}&
    \includegraphics[width=34mm,height=16mm]{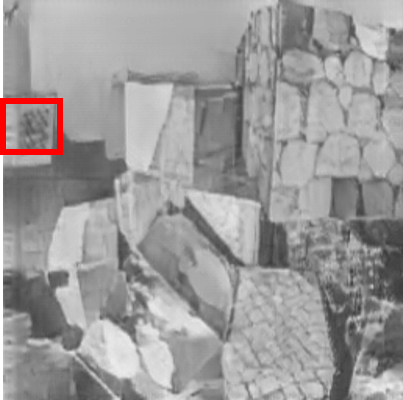}&
    \includegraphics[width=34mm,height=16mm]{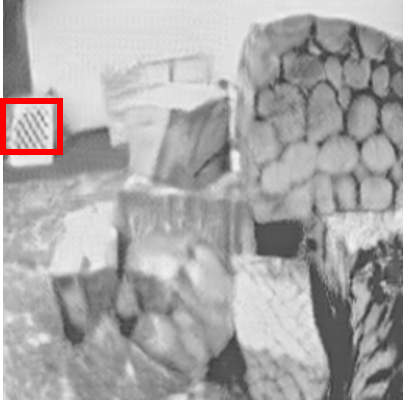}&
    \includegraphics[width=34mm,height=16mm]{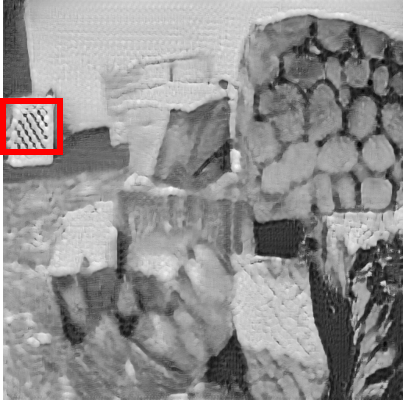}\\
    
    \includegraphics[width=34mm,height=14mm]{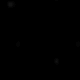}&
    \includegraphics[width=34mm,height=14mm]{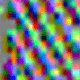}&
    \includegraphics[width=34mm,height=14mm]{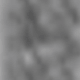}&
    \includegraphics[width=34mm,height=14mm]{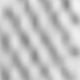}&
    \includegraphics[width=34mm,height=14mm]{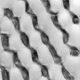}\\

\end{tabular}
\vspace{-13pt}
\captionsetup{font=small}
\caption{Experimental results on HDR effects with event camera dataset \cite{mueggler2017event}. EventSR also works well on reconstructing HDR images. 
} 
\label{fig:hdr_exp}
\end{center}
 \vspace{-16pt}
\end{figure*}

 \begin{figure*}[t!]
\begin{center}
\renewcommand{\tabcolsep}{1pt}
\begin{tabular}{@{}cccccc@{}}
 \small{Stacked Events}& \small{Blurry APS} & \small{Tao \etal\cite{tao2018scale}}&
 \small{Pan \etal\cite{pan2019bringing}} &  \small{Phase.2 Rest.} & \small{Phase.3 SR(x4)}  \\ 
    \includegraphics[width=28.5mm,height=20mm]{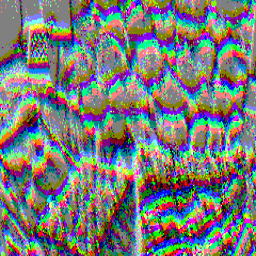}&
    \includegraphics[width=28.5mm,height=20mm]{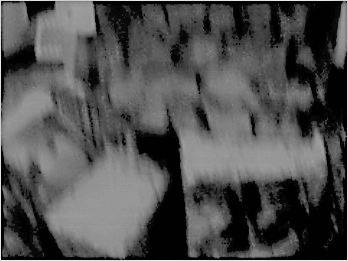}&
    \includegraphics[width=28.5mm,height=20mm]{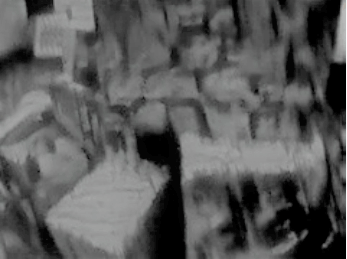} &
    \includegraphics[width=28.5mm,height=20mm]{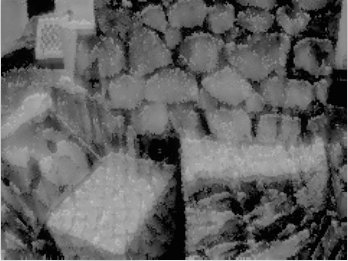}&
    \includegraphics[width=28.5mm,height=20mm]{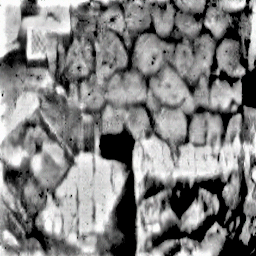}&
    \includegraphics[width=28.5mm,height=20mm]{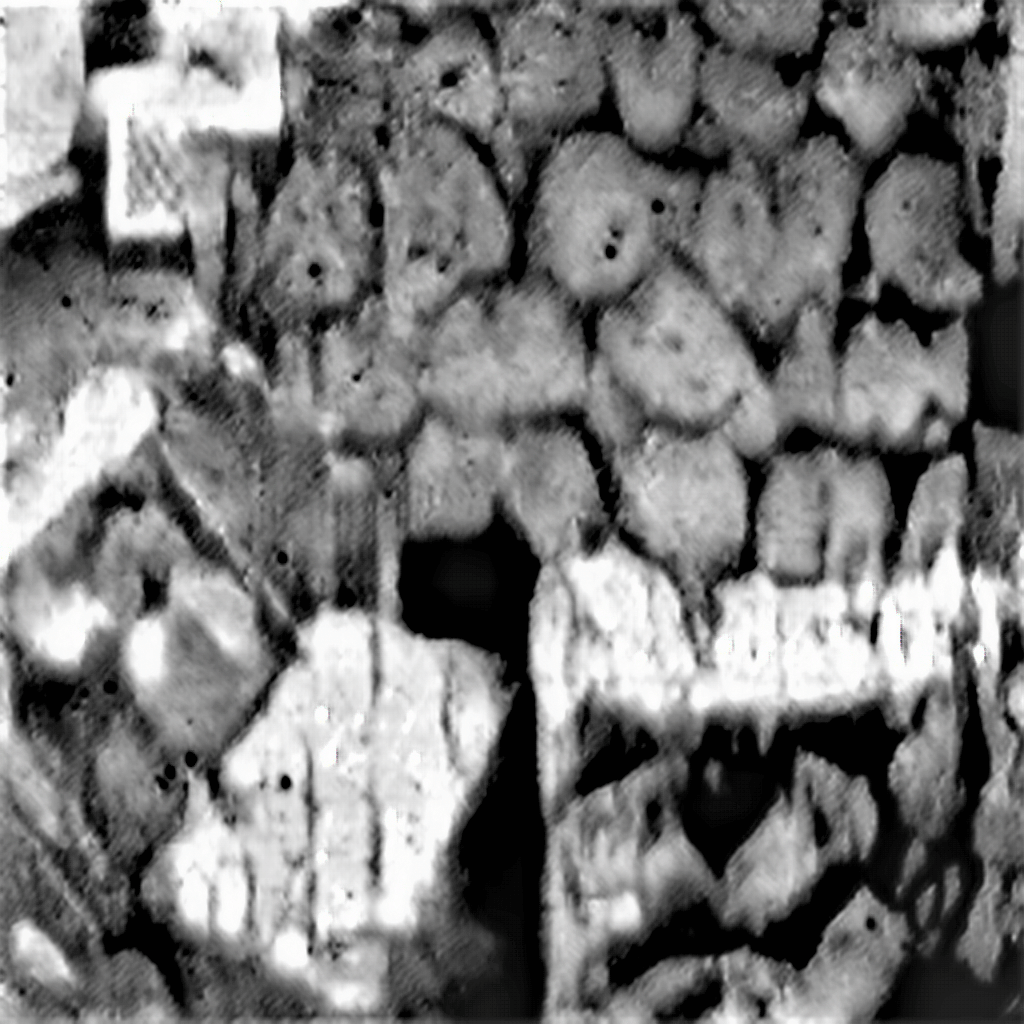}\\
   
\end{tabular}
\vspace{-10pt}
\captionsetup{font=small}
\caption{Qualitative results for motion blur on RW dataset \cite{mueggler2017event}. EventSR achieves better quality than Tao \etal \cite{tao2018scale} and Pan \etal \cite{pan2019bringing}. } 
\label{fig:motionblur_exp}
\end{center}
 \vspace{-20pt}
\end{figure*}

\subsection{Evaluation on real-world dataset}

\noindent\textbf{Evaluation on Ev-RW dataset} We demonstrate EventSR shows more impressive results on real-world data. As mentioned in Sec.~\ref{data_sec}, using real-world data alone is not able to handle the challenges in three phases. We show that training $G_r$ in phase 1 with ESIM-RW dataset and $G_d$ with clean real-world APS images in phase 2 are more advantageous. Note that since we use $\mathcal{L}_{Id}$ in phase 1, we obtain clean RW APS images through $G^{r}$ to get clean RW APS images before training phase 2.   
Figure~\ref{fig:rw_exp} shows the experimental results on Ev-RW dataset.  In phase 1, our method can successfully reconstruct shapes, building, etc, however, the reconstructed images are quite noisy, blurry and unrealistic. In phase 2, the EventSR could restore realistic LR images from events. \emph{This indicates that restoring realistic images from events in phase 2 is a very crucial step for SR image reconstruction.} Although real events are noisy, in phase 3, EventSR can recover the high-frequency structures (\eg lines and textures) and non-blurry SR images from events. The cropped patches in the second and forth rows clearly depict the effectiveness of each stage, in contrast to the APS images. Table.~\ref{table:rw_comp} quantitatively shows EventSR achieves better results than the prior-arts \cite{wang2019event, rebecq2019events, munda2018real, bardow2016simultaneous} regarding phase 1. Our unsupervised method shows lower LPIPS and higher SSIM/FSIM scores than the supervised methods, indicating better reconstruction in phase 1.

 
\noindent\textbf{High dynamic range image} 
In this work, it is apparently shown that events have rich information for HDR image reconstruction, restoration, and super-resolution. Although some parts of the scenes are invisible in APS images due to the low dynamic range, many events do exist in those regions, and EventSR can fully utilize the rich information contained in events to reconstruct HDR SR images. We evaluate the HDR effects \cite{mueggler2017event} with $G_r$ (phase 1) trained with ESIM-RW dataset and with $G_d$ (phase 2) trained using Ev-RW dataset. Figures~\ref{fig:hdr_exp} and \ref{fig:hdr_impressive} show that EventSR successfully reconstructs HDR SR images from pure events. Although in phase 1 the reconstructed images are noisy, blurry and unrealistic, phase 2 recovers the correct shapes and textures of wall poster and boxes. In phase 3, the right structure and informative details are recovered, which can be verified in the cropped patches in second and forth rows.  

\noindent \textbf{Motion deblur} We also demonstrate that EventSR can restore deblurred images from events.  As shown in Fig.~\ref{fig:motionblur_exp}, we visually compare our approach with Tao \etal~\cite{kim2014simultaneous} and Pan \etal~\cite{pan2019bringing} (energy minimization) based on Ev-RW blur effects \cite{mueggler2017event}. Although, the APS image is blurry, our method can restore sharp and realistic images (clear edge, corner, texture, etc) from events in phase 2, which is further enhanced to the non-blurry SR image from events in phase 3. 
\vspace{-10pt}
\subsection{Ablation study}
To validate the effectiveness of the proposed loss functions in EventSR, 
we compare different network structures by removing loss functions selectively.

\noindent\textbf{Remove $F_s$ and $D_s$}  We remove $G_s$ and $D_s$, namely removing $\mathcal{L}_{Adv}^{s}$ and $\mathcal{L}_{Sim}^{s}$. We map the embedded events to clean LR images and  reconstruct SR images using $G_s(G_d(G_r))$. However, without $F_s$ and $D_s$, some noise in the events are mapped to the SR images, affecting visual quality.  

\noindent \textbf{Remove $F_d$ and $D_d$} We also remove $F_d$ and $D_s$ from EventSR, namely, removing $\mathcal{L}_{Adv}^{d}$ and $\mathcal{L}_{Sim}^{d}$. We use EventSR for event to SR image reconstruction, where the forward network is $G_s(G_d(G_r))$ and $F_s$ is the feedback network. We load the pre-trained $G_r$ and $G_d$ and add them to $G_s$. However, without $\mathcal{L}_{Adv}^{d}$ and $\mathcal{L}_{Sim}^d$, the $G_d$ is unable to get the clean images from events.

\noindent\textbf{Remove $F_r$ and $D_r$} We lastly remove $F_r$ and $D_r$, namely $\mathcal{L}_{Adv}^r$ and $\mathcal{L}_{Sim}^r$. However, it shows that $\mathcal{I}^{r}$ is always with undesired artifacts and training is unstable. It is hard to reconstruct SR images from events without these losses. 

\begin{figure}[t!]
\begin{center}
\renewcommand{\tabcolsep}{1pt}
\begin{tabular}{@{}cccc@{}}
 \footnotesize{Stacked events} & \footnotesize{SR(DeblurGAN)} &  \footnotesize{SR (EDSR)} &  \footnotesize{Phase 3 SR(x4)}  \\ 
    \includegraphics[width=20mm,height=12mm]{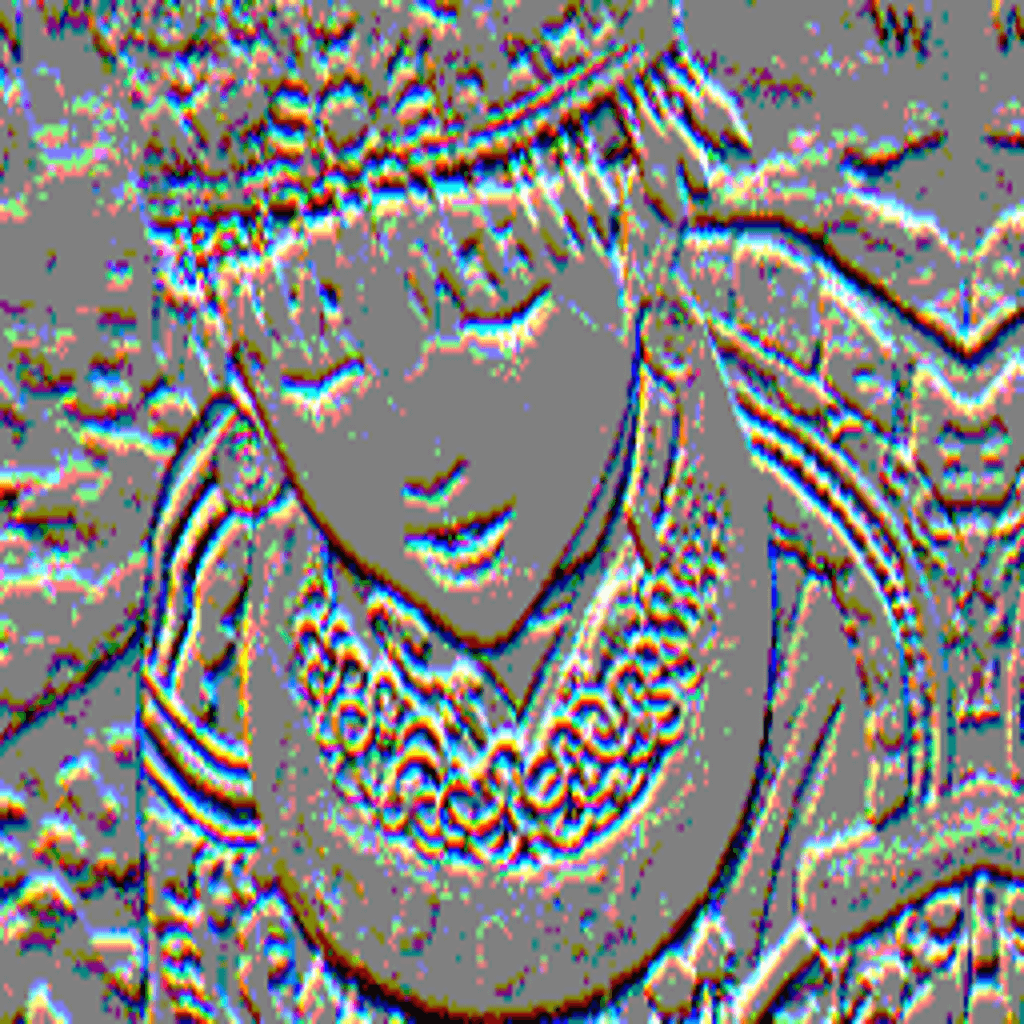}& 
    \includegraphics[width=20mm,height=12mm]{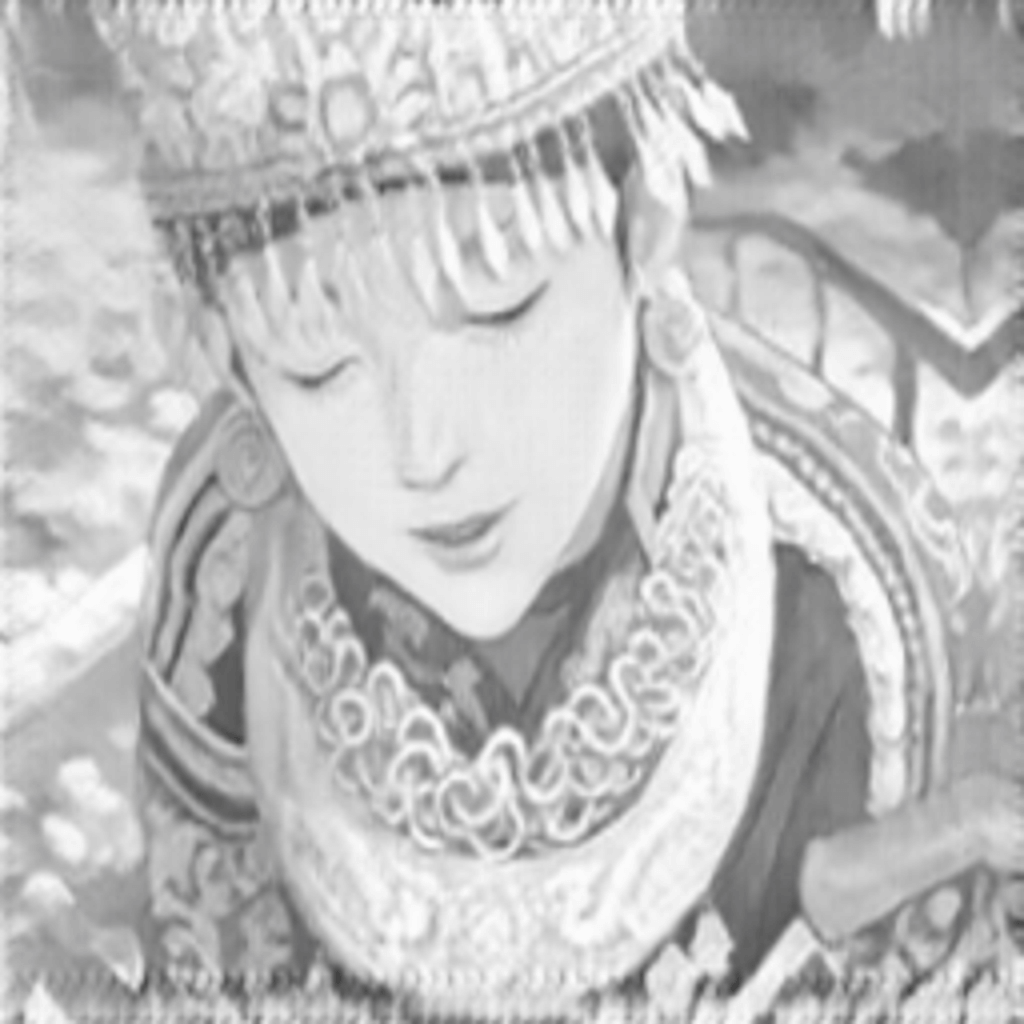}&
    \includegraphics[width=20mm,height=12mm]{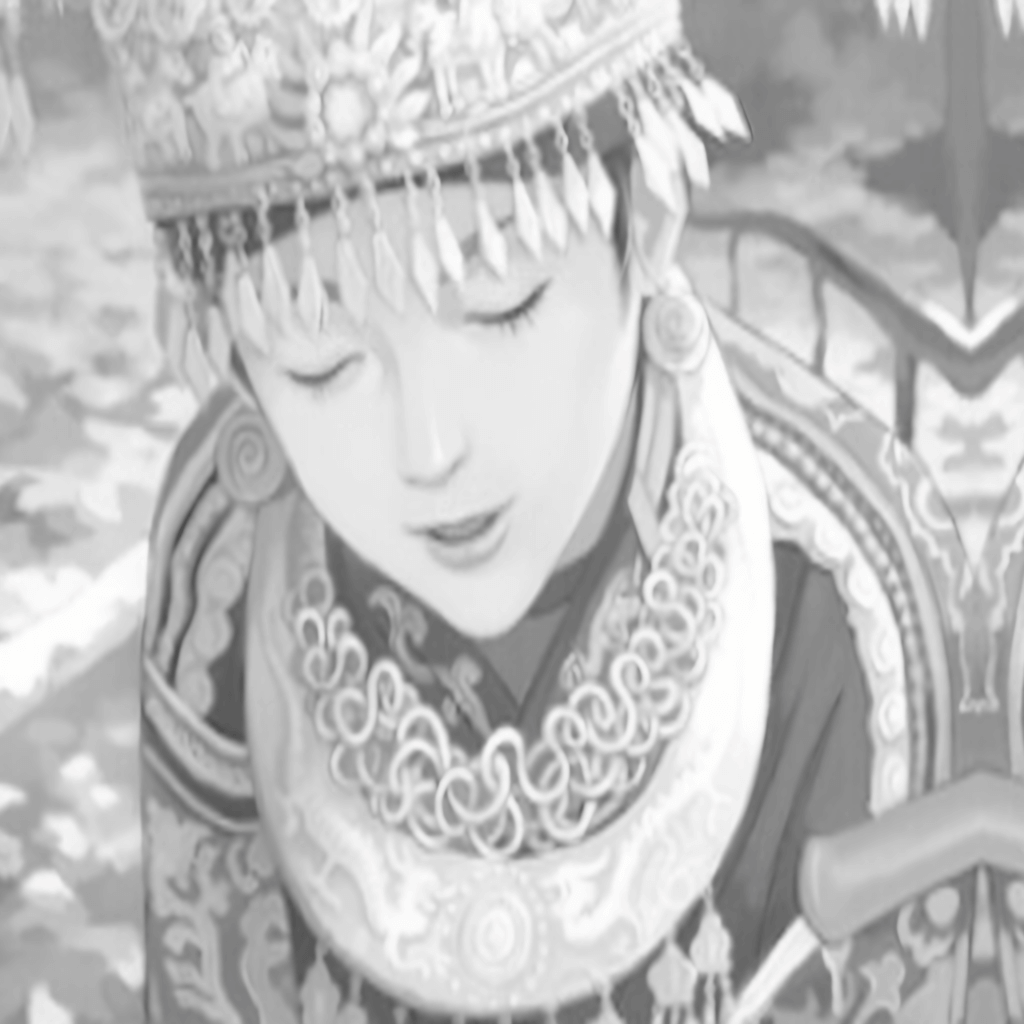}&
    \includegraphics[width=20mm,height=12mm]{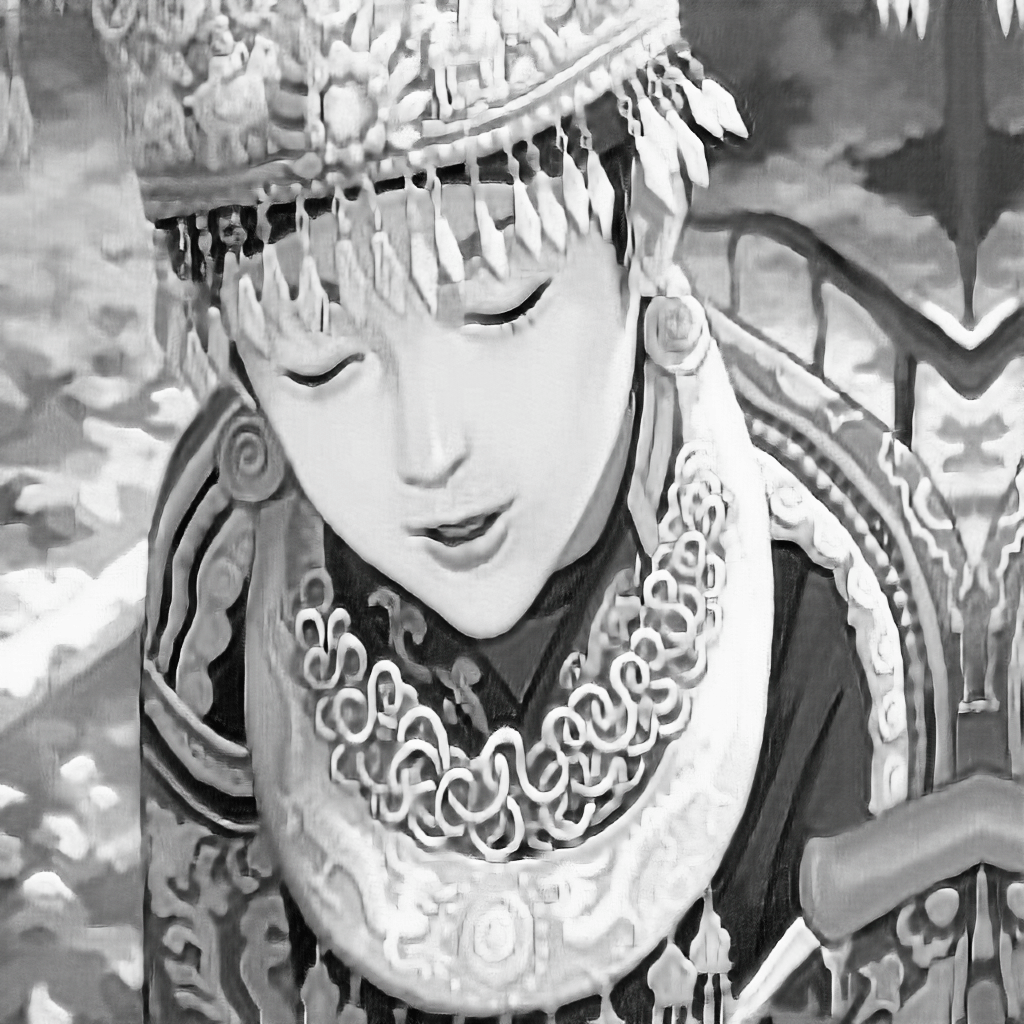}\\
    \includegraphics[width=20mm,height=13mm]{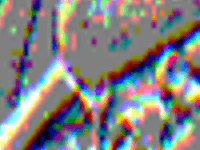}&
    \includegraphics[width=20mm,height=13mm]{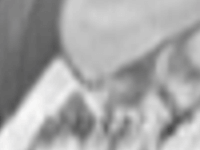}&
    \includegraphics[width=20mm,height=13mm]{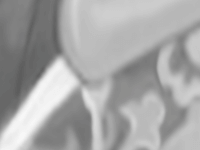}&
    \includegraphics[width=20mm,height=13mm]{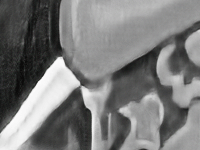}\\
\end{tabular}
\vspace{-12pt}
\captionsetup{font=small}
\caption{ Visual comparison using existing deblur and SR nets.} 
\label{fig:compare_nets}
\end{center}
\vspace{-28pt}
\end{figure}

\vspace{-2pt}
\section{Discussions}
\vspace{-1pt}

\noindent \textbf{Computational cost, observation window, and latency} Our model has around $17$M parameters, which is less than some SOTA SR DNNs, such as EDSR \cite{lim2017enhanced}, RDN \cite{zhang2018residual}, 
thus the training time is comparable with others. The inference time is around $300\sim500$ms on average when using NVIDIA 1080 Ti GPU.  In our experiments, $10K$ events are gathered in $5$ ms time duration on average. Events can be stacked with the fixed observation window as SBT in \cite{wang2019event}.

\noindent \textbf{Using the existing deblur and SR nets}
One might think that the SR results can achieved by directly using the existing deblur and SR networks after phase 1. However, we need to clarify that our method is not just combining networks in a naive way and the phase 2 is not just to deblur but to restore.  
To verify this, we replace phase 2 and 3 with pretrained SOTA deblur and SR networks (\eg DeblurGAN \cite{kupyn2018deblurgan} and EDSR \cite{lim2017enhanced}). As in Fig.~\ref{fig:compare_nets}, one can clearly see that the proposed method (4th column) is superior to such a naive combination.  Without continuously utilized event information, applying existing deblur and SR nets magnifies the noise level and fails to enhance the image quality. 


\noindent \textbf{SR video from events} In this work, we focus on super-resolving HR images from LR events and we do not fully consider the temporal consistency for video.  
However, we will investigate enforcing temporal consistency for super-resolving video from events in our future work.  
\vspace{-2pt}
\section{Conclusion and Future Work}
In this paper, we presented the first and novel framework for event to SR image reconstruction. Facing up with the challenges of no GT images for real-world data in all three phases, we proposed EventSR to learn a mapping from events to SR images in an unsupervised manner. To train EventSR, we made an open dataset including both simulated and real-world scenes. The conjunctive and alternative use of them boosted up the performance.
Experimental results showed that EventSR achieved impressive results even on phase 1 and phase 2, and desirable results in phase 3. 
However, in this work, we have not deeply considered how the forms of event stacks affect the overall performance of EventSR. We will investigate better ways to embed events as input to EventSR as in~\cite{gehrig2019end,tulyakov2019learning} and its potential applications to other tasks in the following work. Besides, we are also aiming to reconstruct SR video from event streams. 
\vspace{-3pt}
\section*{Acknowledgement}
\vspace{-5pt}
This work was supported by the National Research Foundation of Korea(NRF) grant funded by the Korea government(MSIT) (NRF-2018R1A2B3008640) and the Next-Generation Information Computing Development Program through the National Research Foundation of Korea(NRF) funded by the Ministry of Science, ICT (NRF-2017M3C4A7069369).

{\small
\bibliographystyle{ieee_fullname}
\bibliography{eventSR_CVPR2020}
}

\end{document}